\documentclass[11pt,a4paper]{article}
\usepackage[hyperref]{acl2020}
\usepackage{times}
\usepackage{latexsym}

\usepackage{graphicx}  
\usepackage{booktabs} 
\usepackage{multirow}
\usepackage{xcolor}
\usepackage{subcaption}

\usepackage{microtype}

\newcommand{\tablesepremove}{\aboverulesep = 0mm \belowrulesep = 0mm}

\aclfinalcopy 
\def\aclpaperid{1785} 

\title{Expertise Style Transfer: A New Task Towards Better Communication between Experts and Laymen}

\author{Yixin Cao$^1$ \quad Ruihao Shui$^2$$^1$ \quad Liangming Pan$^2$$^1$\\
\textbf{Min-Yen Kan$^1$ \quad Zhiyuan Liu$^3$ \quad Tat-Seng Chua$^1$}\\
$^1$School of Computing, National University of Singapore, Singapore\\
$^2$NUS Graduate School for Integrative Sciences and Engineering\\
$^3$Department of CST, Tsinghua University, Beijing, China\\
{\tt caoyixin2011@gmail.com, \{ruihaoshui,e0272310\}@u.nus.edu} \\
{\tt liuzy@tsinghua.edu.cn, \{kanmy@comp.,dcscts@\}nus.edu.sg}\\
}

\date{}

\begin{document}
\maketitle
\begin{abstract}
  The curse of knowledge can impede communication between experts and laymen. We propose a new task of expertise style transfer and contribute a manually annotated dataset with the goal of alleviating such cognitive biases. Solving this task not only simplifies the professional language, but also improves the accuracy and expertise level of laymen descriptions using simple words. This is a challenging task, unaddressed in previous work, as it requires the models to have expert intelligence in order to modify text with a deep understanding of domain knowledge and structures. We establish the benchmark performance of five state-of-the-art models for style transfer and text simplification. The results demonstrate a significant gap between machine and human performance. We also discuss the challenges of automatic evaluation, to provide insights into future research directions. The dataset is publicly available at \url{https://srhthu.github.io/expertise-style-transfer/}.
\end{abstract}

\section{Introduction}
The curse of knowledge~\cite{camerer1989curse} is a pervasive cognitive bias exhibited across all domains, leading to discrepancies between an expert's advice and a layman's understanding of it~\cite{tan2017internet}. Take medical consultations as an example: patients often find it difficult to understand their doctors' language. On the other hand, it is important for doctors to accurately disclose the exact illness conditions based on patients' simple vocabulary. Misunderstanding may lead to failures in diagnosis and prompt treatment, or even death. How to automatically adjust the expertise level of texts is critical for effective communication.

\begin{figure}[htb]
  \centerline{\includegraphics[width=0.49\textwidth]{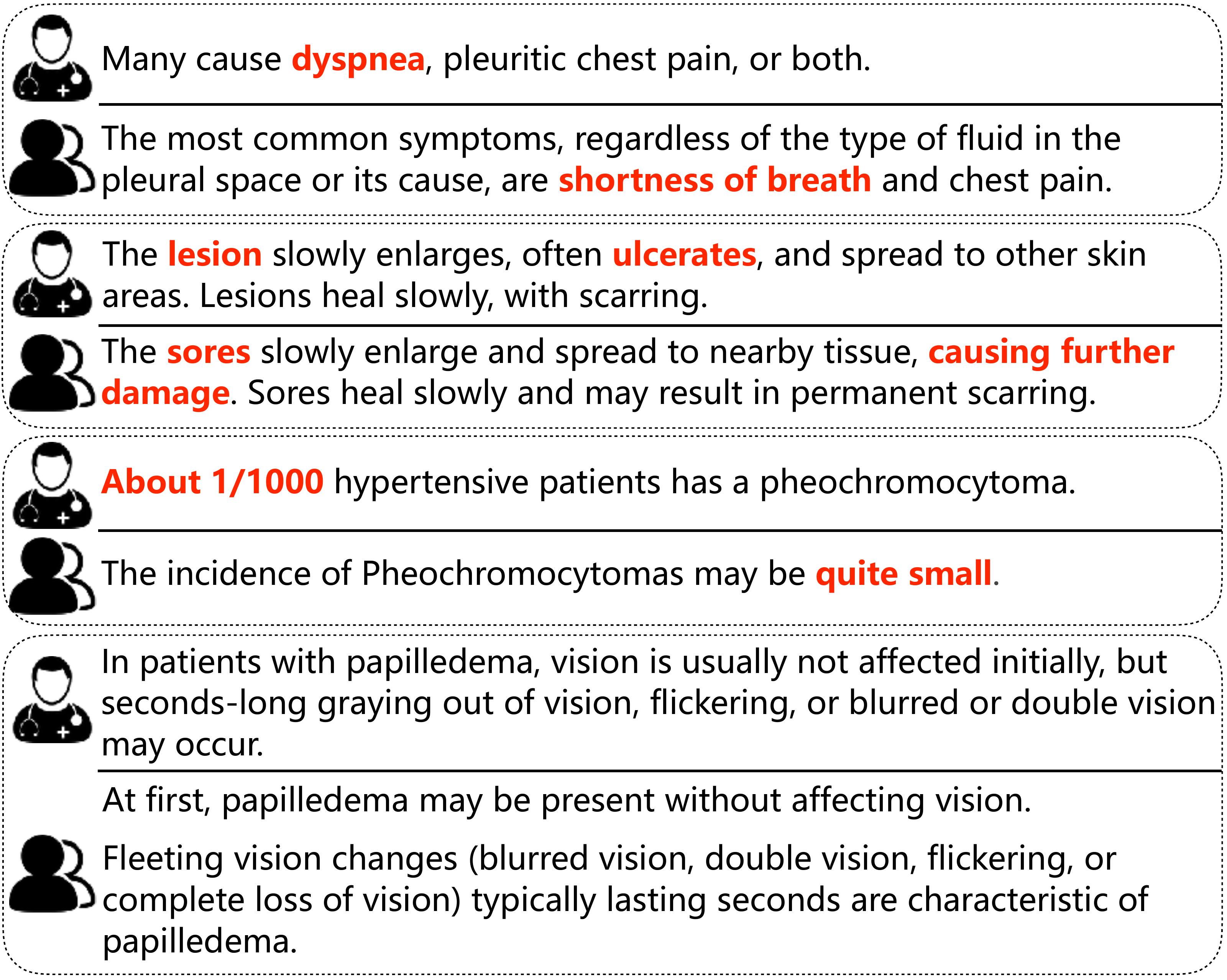}}
	\caption{Examples of Expert Style Transfer. The upper sentences are in expert style while the lower ones are in laymen style. We highlight the knowledge based differences with red bolded font.}
  \label{fig:example}
\end{figure}

In this paper, we propose a new task of text style transfer between expert language and layman language, namely \textbf{Expertise Style Transfer}, and contribute a manually annotated dataset in the medical domain for this task. We show four examples in Figure~\ref{fig:example}, where the upper sentence is for professionals and the lower one is for laymen. On one hand, expertise style transfer aims at improving the readability of a text by reducing the expertise level, such as explaining the complex terminology \textit{dyspnea} in the first example with a simple phrase \textit{shortness of breath}. On the other hand, it also aims to improve the expertise level based on context, so that laymen's expressions can be more accurate and professional. For example, in the second pair, \textit{causing further damage} is not as accurate as \textit{ulcerates}, omitting the important mucous and disintegrative conditions of the sores.


There are two related tasks, but neither serve as suitable prior art. The first is text style transfer (ST), which generates texts with different attributes but with the same content. However, although existing approaches have achieved a great success regarding the attributes of sentiment~\cite{li2018delete} and formality~\cite{rao2018dear} among others, expertise ``styling'' has not been explored yet. Another similar task is Text Simplification (TS), which rewrites a complex sentence with simple structures~\cite{Sulem2018Simple} while constrained by limited vocabulary~\cite{Paetzold2016UnsupervisedLS}. This task can be regarded as similar to our subtask: reducing the expertise level from expert to layman language without considering the opposing direction. However, most existing TS datasets are derived from Wikipedia, and contain numerous noise (misaligned instances) and inadequacies (instances having non-simplified targets)~\cite{xu2015problems,surya2018unsupervised}; in which further detailed discussion can be found in Section~\ref{sec:data_ana}.


In this paper, we construct a manually-annotated dataset for expertise style transfer in medical domain, named MSD, and conduct deep analysis by implementing state-of-the-art (SOTA) TS and ST models. The dataset is derived from human-written medical references, The Merck Manuals\footnote{\url{https://en.wikipedia.org/wiki/The_Merck_Manuals}}, which include two parallel versions of texts, one tailored for consumers and the other for healthcare professionals. For automatic evaluation, we hire doctors to annotate the parallel sentences between the two versions (examples shown in Figure~\ref{fig:example}). Compared with both ST and TS datasets, MSD is more challenging from two aspects:

\textbf{Knowledge Gap.} Domain knowledge is the key factor that influences the expertise level of text, which is also a key difference from conventional styles. We identify two major types of knowledge gaps in MSD: terminology, e.g., \textit{dyspnea} in the first example; and empirical evidence. As shown in the third pair, doctors prefer to use statistics (\textit{About 1/1000}), while laymen do not (\textit{quite small}).

\textbf{Lexical \& Structural Modification.} \citet{Fu2019RethinkingTA} has indicated that most ST models only perform lexical modification, while leaving structures unchanged. Actually, syntactic structures play a significant role in language styles, especially regarding complexity or simplicity~\cite{carroll1999simplifying}. As shown in the last example, a complex sentence can be expressed with several simple sentences by appropriately splitting content. However, available datasets rarely contain such cases.

Our main contributions can be summarized as:

\begin{itemize}
  \item We propose the new task of expertise style transfer, which aims to facilitate communication between experts and laymen.
  \item We contribute a challenging dataset that requires knowledge-aware and structural modification techniques.
  \item We establish benchmark performance and discuss key challenges of datasets, models and evaluation metrics.
\end{itemize}

\section{Related Work}
\label{sec:rw}

\subsection{Text Style Transfer}
Existing ST work has achieved promising results on the styles of sentiment~\cite{hu2017toward,shen2017style}, formality~\cite{rao2018dear}, offensiveness~\cite{Santos2018FightingOL}, politeness~\cite{Sennrich2016ControllingPI}, authorship~\cite{xu2012paraphrasing}, gender and ages~\cite{Prabhumoye2018StyleTT,Lample2019MultipleAttributeTR}, etc. Nevertheless, only a few of them focus on supervised methods due to the limited availability of parallel corpora. \citet{jhamtani2017shakespearizing} extract modern language based Shakespeare's play from the educational site, while \citet{rao2018dear} and \citet{li2018delete} utilize crowdsourcing techniques to rewrite sentences from Yahoo Answers, Yelp and Amazon reviews, which are then utilized for training neural machine translation (NMT) models and evaluation. 

More practically, there is an enthusiasm for unsupervised methods without parallel data. There are three groups. The first group is \textbf{Disentanglement methods} that learn disentangled representations of style and content, and then directly manipulating these latent representations to control style-specific text generation. \citet{shen2017style} propose a cross-aligned autoencoder that learns a shared latent content space between true samples and generated samples through an adversarial classifier. \citet{hu2017toward} utilize neural generative model, Variational Autoencoders (VAEs)~\cite{Kingma2013AutoEncodingVB}, to represent the content as continuous variables with standard Gaussian prior, and reconstruct style vector from the generated samples via an attribute discriminator. To improve the ability of style-specific generation, \citet{fu2018style} utilize multiple generators, which are then extended by a Wasserstein distance regularizer~\cite{zhao2017adversarially}. SHAPED~\cite{Zhang2018SHAPEDSE} learns a shared and several private encoder--decoder frameworks to capture both common and distinguishing features. Some variants further investigate the auxiliary tasks to better preserve contents~\cite{wu2019hierarchical}, or domain adaptation~\cite{Li2019DomainAT}.

Another line of work argues that it is difficult to disentangle style from content. Thus, their main idea is to learn style-specific translations, which are trained using unaligned data based on back-translation~\cite{Zhang2018StyleTA,Prabhumoye2018StyleTT,Lample2019MultipleAttributeTR}, pseudo parallel sentences according to semantic similarity~\cite{Jin2019UnsupervisedTA}, or cyclic reconstruction~\cite{Dai2019StyleTU}, marked with \textbf{Translation methods}.

The third group is \textbf{Manipulation methods}. \citet{li2018delete} first identify the style words by their statistics, then replace them with similar retrieved sentences with a target style. \citet{Xu2018UnpairedST} jointly train the two steps with a neutralization module and a stylization module based on reinforcement learning. For better stylization, \citet{Zhang2018LearningSM} introduce a learned sentiment memory network, while \citet{wu2019hierarchical} utilize hierarchical reinforcement learning.

\subsection{Text Simplification}
Earlier work on text simplification define a sentence as simple, if it has more frequent words, shorter length and fewer syllables per word, etc. This motivates a variety of syntactic rule-based methods, such as reducing sentence length~\cite{chandrasekar1997automatic,vickrey2008sentence}, lexical substitution~\cite{Glavas2015SimplifyingLS,Paetzold2016UnsupervisedLS} or sentence splitting~\cite{woodsend2011learning,Sulem2018Simple}. Another line of work follows the success of machine translation (MT)~\cite{Klein2017OpenNMTOT}, and regards TS as a monolingual translation from complex language to simple language~\cite{zhu2010monolingual,coster2011learning,wubben2012sentence}. \citet{zhang2017sentence} incorporate reinforcement learning into the encoder--decoder framework to encourage three types of simplification rewards concerning language simplicity, relevance and fluency, while \citet{shardlow2019neural} improve the performance of MT models by introducing explanatory synonyms. To alleviate the heavy burden of parallel training corpora, \citet{surya2018unsupervised} propose an unsupervised model via adversarial learning between a shared encoder and separate decoders.

The simplicity of language in the medical domain is particularly important. Terminologies are one of the main obstacles to understanding, and extracting their explanations could be helpful for TS~\cite{shardlow2019neural}. \citet{deleger2008paraphrase} detect paraphrases from comparable medical corpora of specialized and lay texts, and \citet{kloehn2018improving} explore UMLS~\cite{bodenreider2004unified} and WordNet~\cite{miller2009wordnet} with word embedding techniques. Furthermore, \citet{van2019evaluating} directly align sentences from medical terminological articles in Wikipedia and Simple Wikipedia\footnote{\url{https://simple.wikipedia.org/wiki/Main_Page}}, which confines the editors' vocabulary to only 850 basic English words. Then, they refine these aligned sentences by experts towards automatic evaluation. However, the Wikipedia-based dataset is still noisy (with misaligned instances) and inadequate (instances having non-simplified targets) with respect to both model training and testing. Besides, it is usually ignored that the opposite direction of TS --- improving the expertise levels of layman language for accuracy and professionality --- is also critical for better communication.

\subsection{Discussion}

To sum up, both tasks lack parallel data for training and evaluation. This prevents researchers from exploring more advanced models concerning the knowledge gap as well as linguistic modification of lexicons and structures. In this work, we define a more useful and challenging task of expertise style transfer with high-quality parallel sentences for evaluation. Besides, the two communities of ST and TS can shed lights to each other on sentence modification techniques.

\section{Dataset Design}
We describe our dataset construction that comprises three steps: data preprocessing, expert annotation and knowledge incorporation. We then give a detailed analysis.

\subsection{Dataset Construction}
\label{sec:data_build}

\begin{table*}[htp]
  \centering
  \small
	\begin{tabular}{c|p{12cm}}
    \toprule
    \multicolumn{2}{l}{\textbf{Pleural Effusion}, Symptoms} \\ \hline
    Expert & Many cause \emph{dyspnea} {\tiny {\color{red} [C0013404]}}, \emph{pleuritic chest pain} {\tiny {\color{red} [C0008033]}}, or both. \\ \hline
    Laymen & The most common symptoms, regardless of the type of fluid in the pleural space or its cause, are \emph{shortness of breath} {\tiny {\color{red} [C2707305;C3274920]}} and \emph{chest pain} {\tiny {\color{red} [C0008031;C2926613]}}. \\ 
		\bottomrule
  \end{tabular}
  \caption{Examples of parallel annotation in MSD, where the red fonts in brackets denote UMLS concepts.}
	\label{tab:para_example}
\end{table*}

The Merck Manuals, also known as the MSD Manuals, have been the world's most trusted health reference for over 100 years. It covers a wide range of medical topics, and is written through a collaboration between hundreds of medical experts, supervised by independent editors. For each topic, it includes two versions: one tailored for consumers and the other for professionals.

\begin{figure}[htb]
  \centerline{\includegraphics[width=\linewidth]{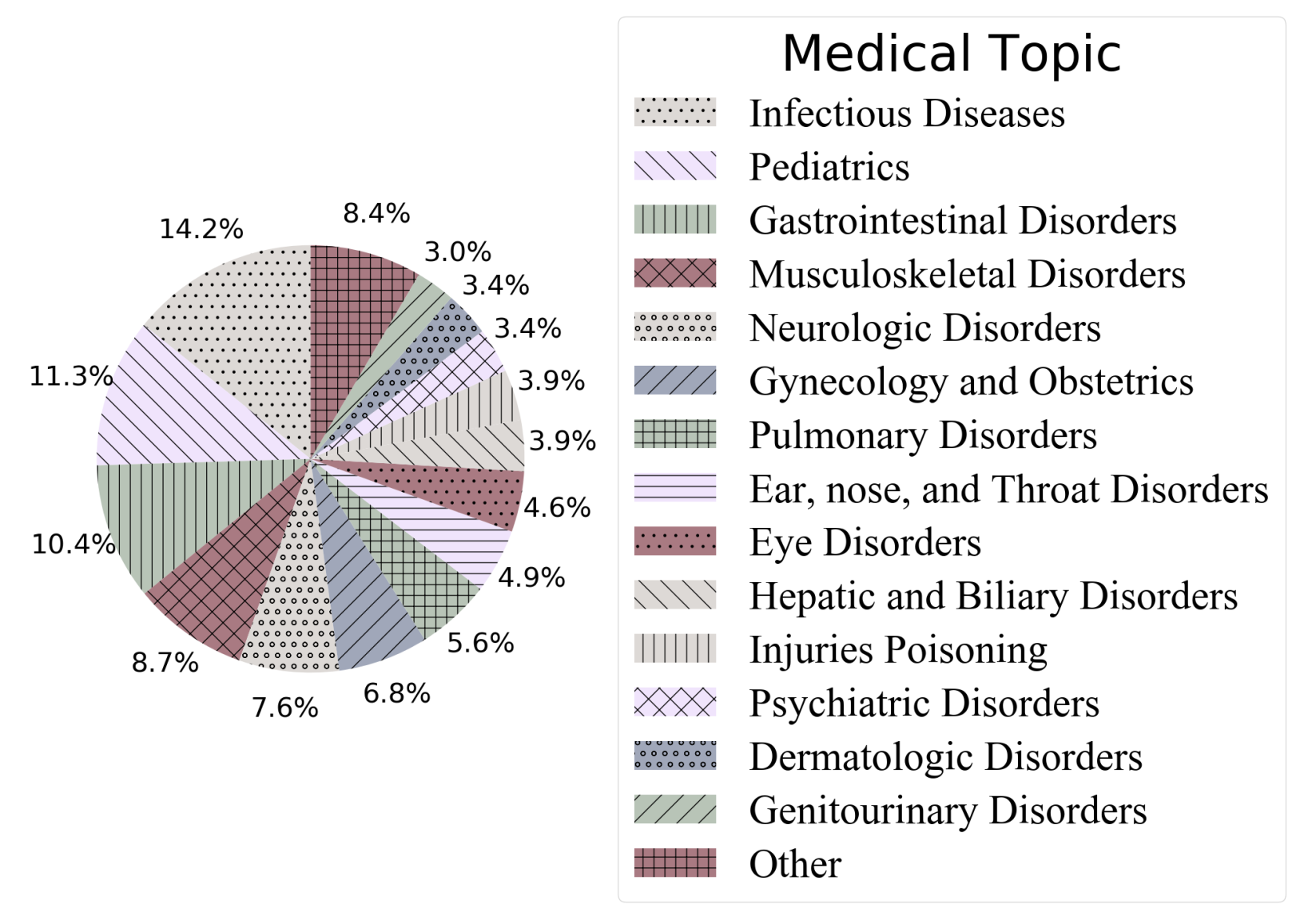}}
  \caption{Distribution of dataset based on topics}
    \label{fig:topic}
\end{figure}

\textbf{Step 1: Data Preprocessing.}
Although the two versions of documents refer to the same topic, they are not aligned, as each document is written independently. We first collect the raw texts from the MSD website\footnote{\url{https://www.msdmanuals.com/}}, and obtain 2601 professional and 2487 consumer documents with 1185 internal links among them. We then split each document into sentences, with the resultant distribution of medical topics as shown in Figure~\ref{fig:topic}. Finally, to alleviate the annotation burden, we find possible parallel groups of sentences by matching their document titles and subsection titles, which denote medical PCIO elements, such as the Diagnosis and Symptoms. Specifically, we first disambiguate the internal links by matching the document title and its accompanied ICD-9 code. Then, we manually align medical PCIO elements in the two versions to provide fine-grained internal links. For example, all sentences for \textit{Atherosclerosis.Symptoms} in the professional MSD may be aligned with those for \textit{Atherosclerosis.Signs} in the consumer MSD. We thus obtain 2551 linked sentence groups as candidates for experts to annotate. Each group contains 10.40 and 11.33 sentences on average for the professional and consumer versions, respectively. We then randomly sample 1000 linked groups for expert annotations in the next section\footnote{The testing size is consistent with other ST datasets, and the rest of groups will be annotated for a larger dataset in the future.}.

\begin{table*}[htbp]
  \tablesepremove
  \small
  \centering
    \begin{tabular}{|r|ccc|ccc|ccc|}
      \toprule
      \multirow{2}[4]{*}{\textbf{Metric}} & \multicolumn{3}{c|}{\textbf{MSD Train}} &
      \multicolumn{3}{c|}{\textbf{MSD Test}} & 
      \multicolumn{3}{c|}{\textbf{SimpWiki}} \\ \cline{2-10}
       & \textbf{Expert} & \textbf{Layman} & \textbf{Ratio} & \textbf{Expert} & \textbf{Layman} & \textbf{Ratio} & \textbf{Expert} & \textbf{Layman} & \textbf{Ratio} \\
      \midrule
      \#Annotation  & 0 & 0 & - & 675 & 675 & - & 2,267 & 2,267 & - \\
      \#Sentence  & 130,349 & 114,674 & - & 930 & 1,047 & 1.13 & 2,326 & 2,307 & 0.99 \\
      \#Vocabulary     & 60,627 & 37,348 & 0.62 & 4,117 & 3,350 & 0.81 & 10,411 & 8,823 & 0.85 \\
      \#Concept Vocabulary & 24,153 & 15,060 & 0.62 & 1,865 & 1,520 & 0.81 & 2,899 & 2,458 & 0.85 \\ \hline
      FleshKincaid      & 12.61 & 9.97 & 0.79 & 12.05 & 9.53 & 0.79 & 12.10 & 9.63 & 0.80 \\
      Gunning           & 18.43 & 15.29 & 0.83 & 17.89 & 15.07 & 0.84 & 17.66 & 14.86 & 0.84 \\
      Coleman           & 12.66 & 10.41 & 0.82 & 12.26 & 9.74 & 0.79 & 10.89 & 9.70 & 0.89 \\
      Avg. Readability & 14.57 & 11.89 & 0.81 & 14.07 & 11.45 & 0.81 & 13.55 & 11.40 & 0.84 \\
      \bottomrule
    \end{tabular}%
  \caption{Statistics of MSD and SimpWiki. One annotation may contain multiple sentences, and MSD Train has no parallel annotations due to expensive expert cost. The ratio of layman to expert according to each metric denotes the gap between the two styles, and a higher value implies smaller differences except that for \#Sentence.}
  \label{tab:stat_comp}
\end{table*}%

\begin{figure}[htb]
  \centerline{\includegraphics[width=0.95\linewidth]{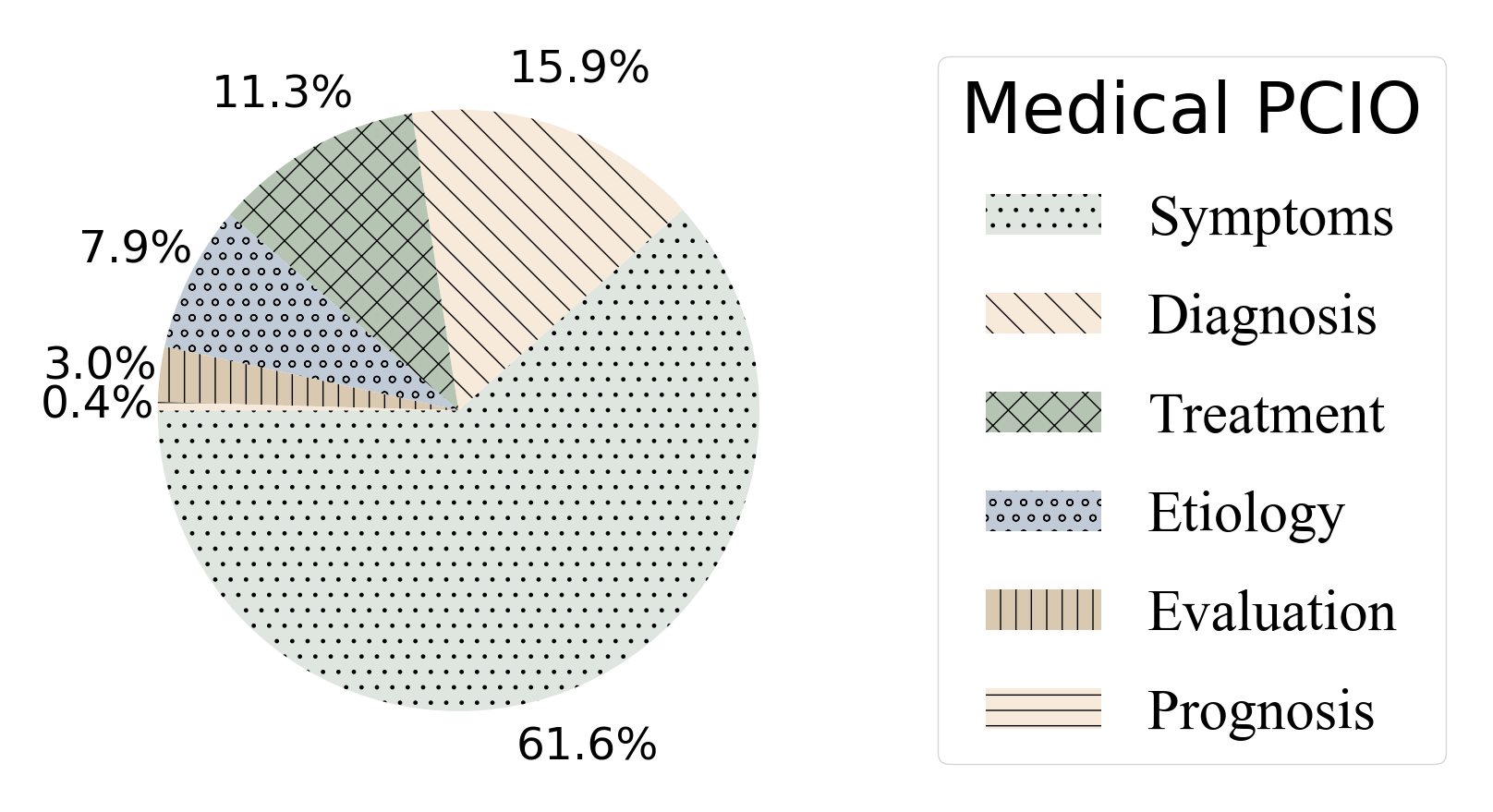}}
  \caption{Distribution of testing set based on PCIO.}
  \label{fig:type}
\end{figure}

\textbf{Step 2: Expert Annotation.}
Given the aligned groups of sentences in professional and consumer MSD, we develop an annotation platform to facilitate expert annotations. We hire three doctors to select sentences from each version of group to annotate pairs of sentences that have the same meaning but are written in different styles. The hired doctors are formally medically trained, and are qualified to understand the semantics of the medical texts. To avoid subjective judgments in the annotations, they are not allowed to change the content. Particularly, the doctors are Chinese who also know English as a second language. Thus, we provide the English content accompanied with a Chinese translation as assistance, which helps to increase the annotation speed while ensuring quality. We also conduct verification on each pair of parallel sentences with the help of another doctor. Note that each pairing may contain multiple professional and consumer sentences; i.e., multiple alignment is possible, the alignments are not necessarily one-to-one. The strict procedure also discards many aligned groups, leading to 675 annotations for testing, with distribution of medical PCIO elements as shown in Figure~\ref{fig:type}.

\textbf{Step 3: Knowledge Incorporation.}
To facilitate knowledge-aware analysis, we can utilize information extraction techniques~\cite{cao2018neural,cao2019low} to identify medical concepts in each sentence. Here, we use QuickUMLS~\cite{soldaini2016quickumls} to automatically link entity mentions to Unified Medical Language System (UMLS)~\cite{bodenreider2004unified}. Note that each mention may refer to multiple concepts, each for which we align to the highest ranked one. As shown in Table~\ref{tab:para_example}, the mention \textit{dyspnea} is linked to concept \textit{C0013404}.

Through this three step process, we obtain a large set of (non-parallel) training sentences in each style, and a small set of parallel sentences for evaluation. The detailed statistics as compared with other datasets can be found in Table~\ref{tab:stat_comp} and Table~\ref{tab:para_comp}. 

\subsection{Dataset Analysis}
\label{sec:data_ana}

Let us compare our MSD dataset against both publicly available ST and TS datasets. \textbf{SimpWiki}~\cite{van2019evaluating} is a TS dataset derived from the linked articles between Simple Wikipedia and Normal Wikipedia. It focuses on the medical domain and extracts parallel sentences automatically by computing their BLEU scores. \textbf{GYAFC}~\cite{rao2018dear} is the largest ST dataset on formality in the domains of \textit{Entertainment \& Music} (E\&M) and \textit{Family \& Relationships} (F\&R) from Yahoo Answers. It contains more than 50,000 training sentences (non-parallel) for each domain, and over 1,000 parallel sentences for testing, obtained by rewriting informal answers via Amazon Mechanical Turk. \textbf{Yelp} and \textbf{Amazon}~\cite{li2018delete} are sentiment ST datasets by rewriting reviews based on crowdsourcing. They both contain over 270k training sentences (non-parallel) and 500 parallel sentences for evaluation. \textbf{Authorship}~\cite{xu2012paraphrasing} aims at transferring styles between modern English and Shakespearean English. It contains 18,395 sentences for training (non-parallel) and 1,462 sentence pairs for testing.

\noindent\textbf{Dataset Statistics}

Table~\ref{tab:stat_comp} presents the statistics of expertise and layman sentences in our dataset as well as SimpWiki. We split the sentences using NLTK, and compute the ratio of layman to expert in each metric to denote the gap between the two styles (a lower value implies a smaller gap expect that for \#Sentence). Three standard readability indices are used to evaluate the simplicity levels: FleshKincaid~\cite{kincaid1975derivation}, Gunning~\cite{gunning1968technique} and Coleman~\cite{coleman1975computer}. The lower the indices are, the simpler the sentence is. Note that SimpWiki does not provide a train/test split, and thus we randomly sample 350 sentence pairs for evaluation. We follow the same strategy in our experiments.

Compared with SimpWiki, we can see that: \textbf{(1)} MSD evaluates the structure modifications. As the layman language usually requires more simple sentences to express the same meaning as in the expert language, each expert sentence in MSD Test refers to 1.13 layman sentences on average, while the number in SimpWiki is only 0.99. \textbf{(2)} MSD is more distinct between the two styles, which is critical for style transfer. This is markedly demonstrated by the larger difference between their (concepts) vocabulary sizes (0.62/0.81 vs. 0.85 in ratio of layman to expert), and between the readability indices (0.81/0.81 vs. 0.84 on average). \textbf{(3)} we have more complex professional sentences in expert language (14.57/14.07 vs. 13.55 in the three readability indices on average) but comparatively simple sentences in laymen language (11.89/11.45 vs. 11.40). This is intuitive because both versions of Wikipedia are written by crowdsourcing editors, and MSD is written by experts in medical domain.

\noindent\textbf{Quality of Parallel Sentences}

One of the main concerns in ST is the limitations of parallel sentences towards automatic evaluation. On one hand, assuming that the parallel sentences have the same meaning, many datasets find the aligned sentences to have higher string overlap (as measured by BLEU). On the other hand, the two sentences should have different styles, and may vary a lot in expressions: and thus leading to a lower BLEU. Hence how to build a testing dataset that considers both criteria is critical. We analyze the quality of testing sentence pairs in each dataset.

\begin{table}[htbp]
  \tablesepremove
  \small
  \centering
  \begin{tabular}{|r|cc|c|} \toprule
    \textbf{Datasets} & \textbf{BLEU} & \textbf{ED} & \textbf{Task} \\ \midrule
    GYAFC{\tiny E\&M} & 16.22 & 28.53 & ST  \\
    GYAFC{\tiny F\&R} & 16.95 & 29.35 & ST \\
    Yelp              & 24.76 & 22.20 & ST \\
    Amazon            & 44.52  & 19.75 & ST \\
    Authorship        & 19.43 & 36.70 & ST \\ 
    SimpWiki          & 49.98 & 64.16 & TS \\ \hline
    MSD               & 14.01 & 139.73 & ST \& TS \\ \bottomrule
  \end{tabular}%
\caption{BLEU (4-gram) and edit distance (ED ) scores between parallel sentences. Concept words are masked for ED computation~\cite{Fu2019RethinkingTA}. Higher BLEUs imply two more similar sentences, while higher edit distances imply more heterogeneous structures.}
\label{tab:para_comp}
\end{table}%

Table~\ref{tab:para_comp} presents the BLEU and edit distance (ED for short) scores. Note that each pair of parallel sentences is verified to convey the same meaning during annotation. We see that: \textbf{(1)} MSD has the lowest BLEU and highest ED. This implies that MSD is very challenging that requires both lexical and structural modifications. \textbf{(2)} TS datasets reflect more structural differences (with higher ED values) as compared to ST datasets. This means that TS datasets concerning the nature of language complexity (simplicity) are more complex to transfer.

\section{Experiments}

We reimplement five SOTA models from prior TS and ST studies on both MSD and SimpWiki datasets. A further ablation study gives a detailed analysis of the knowledge and structure impacts, and highlights the challenges of existing metrics.

\begin{table*}[htbp]
  \tablesepremove
  \small
  \centering
  \resizebox{\linewidth}{!}{%
  \begin{tabular}{c|c|cc|ccc|cc|ccc}
      \toprule
       & \textbf{Dataset} & \multicolumn{5}{c|}{\textbf{SimpWiki}} & \multicolumn{5}{c}{\textbf{MSD}} \\ \cline{2-12}

       & \textbf{Metrics} & \textbf{Acc} & \textbf{PPL} & \textbf{self-BLEU} & \textbf{ref-BLEU} & \textbf{human} & \textbf{Acc} & \textbf{PPL} & \textbf{self-BLEU} & \textbf{ref-BLEU} & \textbf{human} \\

      \midrule
      \multirow{ 6}{*}{E2L} & OpenNMT+PT & 44.29 & 6.88 & 93.38 & 50.16 & 3.99 & 16.00 & 5.95 & 59.89 & 9.91 & 3.53 \\
       & UNTS & 55.14 & 22.06 & 44.80 & 31.11 & 2.96 & 22.07 & 24.62 & 20.49 & 3.94 & 2.66 \\ \cline{2-12}
       & ControlledGen & 46.57 & 21.76 & 58.33 & 29.21 & 2.78 & 11.7 & 5.77 & 88.61 & 13.13 & 3.78 \\ 
       & DeleteAndRetrieve & 38.29 & 5.10 & 0.74 & 0.65 & 1.19 & 74.67 & 3.92 & 6.66 & 2.95 & 2.28 \\ 
       & StyleTransformer & 42.86 & 13.30 & 74.93 & 41.72 & 3.43 & 40.89 & 12.12 & 53.66 & 10.09 & 3.31 \\ \cline{2-12}
       & Gold & 62.00 & 8.36 & 49.91 & - & - & 93.63 & 4.23 & 14.00 & - & - \\

      \midrule\midrule

      \multirow{ 4}{*}{L2E} & ControlledGen & 40.29 & 33.09 & 63.21 & 29.40 & 2.83 & 6.22 & 5.02 & 93.05 & 13.77 & 3.92 \\ 
       & DeleteAndRetrieve & 37.43 & 5.72 & 0.00 & 0.41 & 1.14 & 64.59 & 4.23 & 6.65 & 2.77 & 2.33 \\ 
       & StyleTransformer & 39.43 & 12.91 & 77.94 & 36.63 & 3.36 & 49.33 & 8.04 & 48.36 & 10.57 & 3.25 \\ \cline{2-12}
       & Gold & 58.86 & 6.93 & 50.13 & - & - & 88.15 & 4.34 & 14.01 & - & - \\
      \bottomrule
  \end{tabular}%
  }
  \caption{Overall performance based on style transfer evaluation metrics from expertise to laymen language (marked as E2L) and in the opposite direction (L2E). Gold denotes human references.}
  \label{tab:perf}%
\end{table*}%

\subsection{Baselines}
We choose the following methods to establish benchmark performance on the two datasets on expertise style transfer, because they: (1) achieve SOTA performance in their fields; (2) are typical methods (as grouped in Section~\ref{sec:rw}); and (3) release codes for reimplementation. 

The TS models\footnote{We only report TS models for expertise to laymen language, since they do not claim the opposite direction.} selected are: (1) Supervised model \textbf{OpenNMT+PT} that incorporates a phrase table into OpenNMT~\cite{Klein2017OpenNMTOT}, which provides guidance for replacing complex words with their simple synonym~\cite{shardlow2019neural}; and (2) Unsupervised model \textbf{UNTS} that utilizes adversarial learning~\cite{surya2018unsupervised}.

The models for ST task selected are: (1) Disentanglement method \textbf{ControlledGen}~\cite{hu2017toward} that utilizes VAEs to learn content representations following a Gaussian prior, and reconstructs a style vector via a discriminator; (2) Manipulation method \textbf{DeleteAndRetrieve}~\cite{li2018delete} that first identifies style words with a statistical method, then replaces them with target style words derived from given corpus; and (3) Translation method \textbf{StyleTransformer}~\cite{Dai2019StyleTU} that uses cyclic reconstruction to learn content and style vectors without parallel data.

\subsection{Training Details}
We use the pre-trained OpenNMT+PT model released by the authors\footnote{\url{https://github.com/senisioi/NeuralTextSimplification/}}. Other models are trained using MSD and SimpWiki training data. We leave 20\% of the training data for validation. The training settings follow the standard best practice; where all models are trained using Adam~\cite{DBLP:journals/corr/KingmaB14} with mini-batch size $32$, and the hyper-parameters are tuned on the validation set. We set the shared parameters the same for baseline models: the maximum sequence length is 100, the word embeddings are initialized with 300-dimensional GloVe~\cite{DBLP:conf/emnlp/PenningtonSM14}, learning rate is set to $0.001$, and adaptive learning rate decay is applied. We adopt early stopping and dropout rate is set to $0.5$ for both encoder and decoder.

\subsection{Evaluation Metrics}
Following~\citet{Dai2019StyleTU}, we make an automatic evaluation on three aspects:

\textbf{Style Accuracy} (marked as Acc) aims to measure how accurate the model controls sentence style. We train two classifiers on the training set of each dataset using fasttext~\cite{joulin2017bag}.

\textbf{Fluency} (marked as PPL) is usually measured by the perplexity of the transferred sentence. We fine-tune the state-of-the-art pretrained language model, Bert~\cite{devlin2019bert}, on the training set of each dataset for each style.

\textbf{Content Similarity} measures how much content is preserved during style transfer. We calculate 4-gram BLEU~\cite{papineni2002bleu} between model outputs and inputs (marked as self-BLEU), and between outputs and gold human references (marked as ref-BLEU).

Automatic metrics for content similarity are arguably  unreliable, since the original inputs usually achieve the highest scores~\cite{Fu2019RethinkingTA}. We thus also conduct human evaluation. To evaluate over the entire test set, only layman annotators are involved, but we ensure that the layman style sentences are accompanied as references to assist understanding. Each annotator is asked to rate the model output given both input and gold references. The rating ranges from 1 to 5, where higher values indicate that more semantic content is preserved.

\textbf{Text Simplification Measurement.} The above metrics may not perform well regarding language simplicity~\cite{sulem2018bleu}. So, we also utilize a TS evaluation metrics: SARI~\cite{xu2016optimizing}. It compares the n-grams of the outputs against those of the input and human references, and considers the added, deleted and kept words by the system.

\begin{figure*}[htbp]
  \centering
  \begin{subfigure}[t]{0.33\textwidth}
    \centerline{\includegraphics[width=\linewidth]{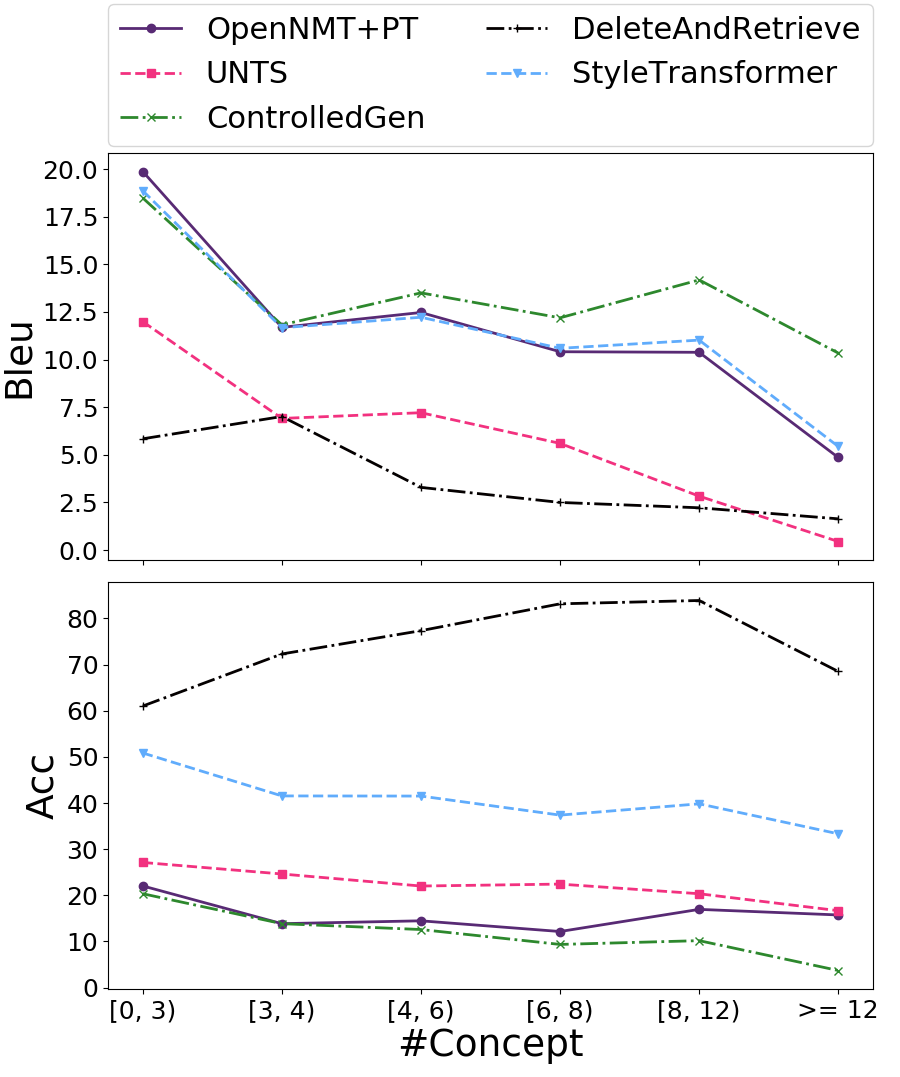}}
  \caption{Impact of concepts.}
	\label{fig:abl_concept}
\end{subfigure}
\hfill
\begin{subfigure}[t]{0.32\textwidth}
  \centerline{\includegraphics[width=\linewidth]{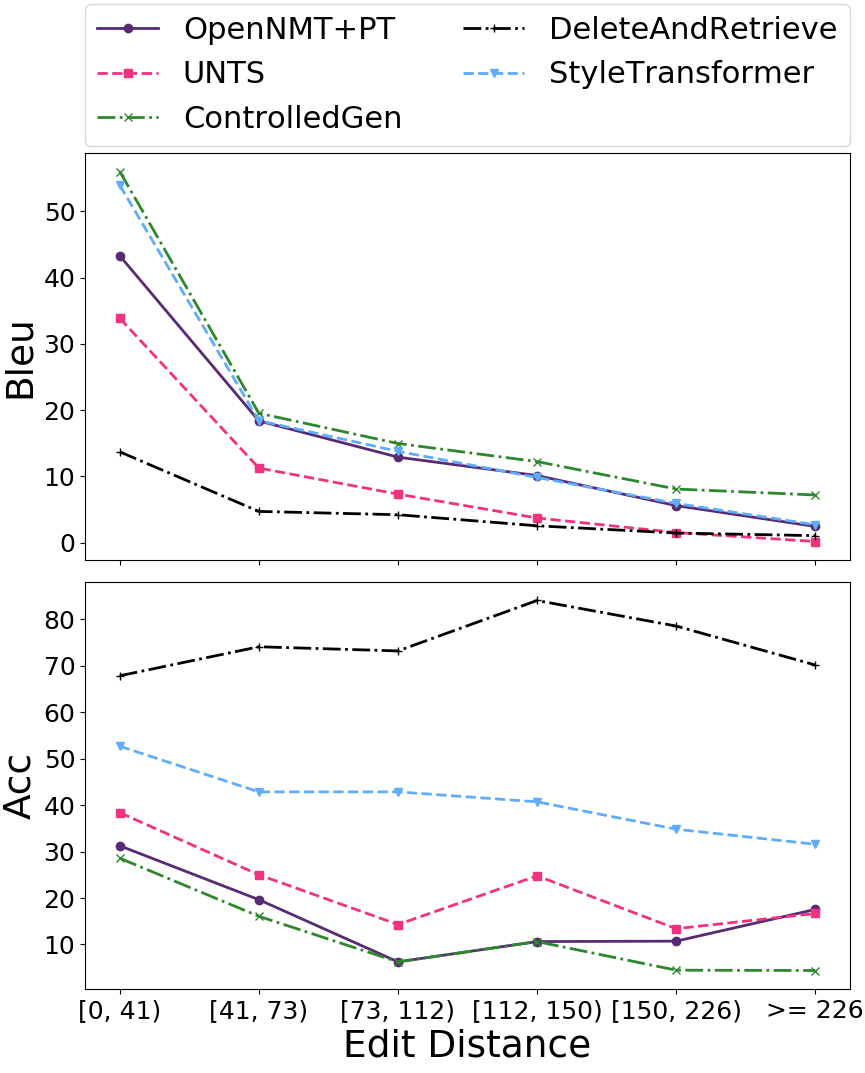}}
\caption{Impact of structure differences.}
\label{fig:abl_ed}
\end{subfigure}
\hfill
\begin{subfigure}[t]{0.325\textwidth}
  \centerline{\includegraphics[width=\linewidth]{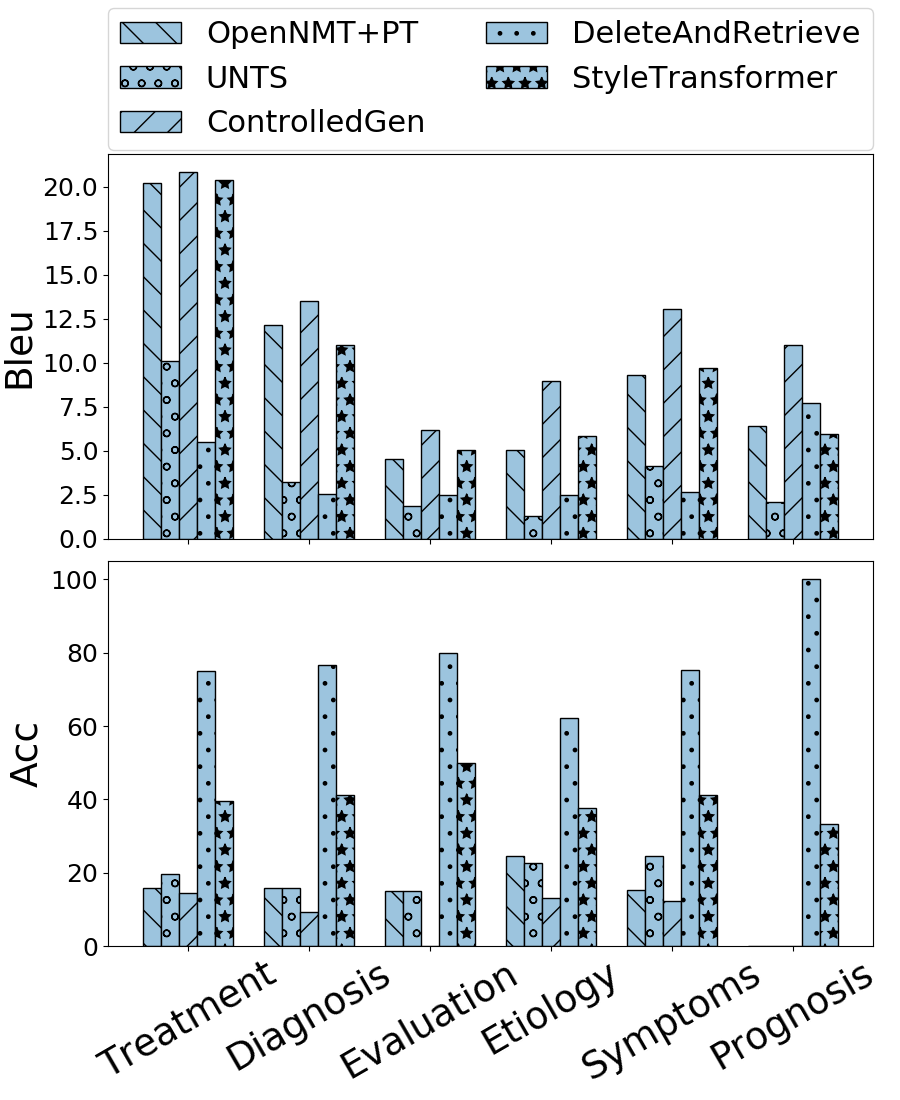}}
\caption{Performance on different PCIO.}
\label{fig:abl_type}
\end{subfigure}
\vspace{-0.2cm}
\caption{Curves of BLEU and style accuracy, where the x-axis denotes the number of concepts per sentence, edit distance between parallel sentences, and different PCIO elements, respectively.}
\label{fig:abl}
\end{figure*}%

\subsection{Overall Performance}
Table~\ref{tab:perf} present the overall performance. Since each pair of parallel sentences has been verified during annotation, we did not report human scores to avoid repeated evaluations. We can see that: \textbf{(1)} Parallel sentences in MSD have higher quality than SimpWiki, because our gold references are more fluent (4.29 vs. 7.65 in perplexity on average) and more discriminable (91\% vs. 60\% on average style accuracy). \textbf{(2)} The transfer for L2E is more difficult (except in content similarity) than that for E2L: 39.55\% vs. 42.50\% in Acc on average, 11.50 vs. 10.33 in PPL on average and 2.80 vs. 2.63 in human ratings on average. This is because the increase in expertise levels requires more contexts and knowledge, and is harder than simplification. \textbf{(3)} TS models perform similarly with ST models. Besides, supervised model OpenNMT+PT outperforms the unsupervised UNTS in fluency and content similarity due to the additional supervision signals. On the other hand, UNTS achieves higher Acc since it utilizes more non-parallel training data. \textbf{(4)} The style accuracy is the reverse to content similarity, making it more challenging to propose a comprehensive evaluation metric that can balance the two opposite directions. In terms of content similarity, even if both self-BLEU and ref-BLEU show a strong correlation with human ratings (over 0.98 Pearson coefficient with p-value$<0.0001$), the higher scores of ControlledGen cannot demonstrate its superior performance, as it actually makes little modifications to styles. Instead, DeleteAndRetrieve, presents a strong ability to control styles (70\% on average in Acc on MSD), but hardly preserves the contents. Style Transformer performs more stably. 

Next, we discuss key factors of MSD. We take the E2L as the exemplar for discussion, as we have observed similar results for the opposing direction.

\subsection{Impact of Concepts}
Figure~\ref{fig:abl_concept} shows the performance curves of BLEU and style accuracy. We choose the concept range to ensure they contain similar number of sentences. Along with the increasing number of concepts, we can see a downward BLEU trend.  This is  because it becomes more difficult to preserve content when the sentence is more professional. As for style accuracy, DeleteAndRetrieve achieves the peak around [8,12) concepts, while the performance of other models drops gradually. Clearly, a lower number of concepts benefit the model for better understanding the sentences due to their correlated semantics, but a larger number of concepts requires knowledge-aware text understanding.

\subsection{Impact of Structures}

Figure~\ref{fig:abl_ed} presents the performance curves regarding the structure differences, where the edit distance is computed as mentioned in Section~\ref{sec:data_ana}.  Higher score denotes more heterogeneous structures. We see a similar trend with the curves of concepts. That is, existing models perform well in simple cases (fewer concepts and less structural differences), but becomes worse if the language is complex. We doubt that the encoder in each model is able to understand the domain-specific language sufficient well without considering knowledge. We thus propose a simple variant of ControlledGen by introducing terminology definitions, and observe some interesting findings in Section~\ref{sec:disc}.

\subsection{Performance on Medical PCIO}
The style of medical PCIO elements (e.g., symptoms) are slightly different. We separately evaluate each model and present the results in Figure~\ref{fig:abl_type}. Style accuracy remains similar among these medical PCIO elements, but there are significant differences among the models in their performance for preserving content. Specifically, models perform well for those sentences about \textit{treatment}, but perform poorly for \textit{evaluation}, because this type of sentences usually involve many rare terms, challenging understanding.

\subsection{Performance using Simplification Metrics}

\begin{table}[htp]
  \centering
  \small
	\begin{tabular}{c|cc|cc}
    \toprule
    \textbf{Dataset} & \multicolumn{2}{c|}{\textbf{SimpWiki}} & \multicolumn{2}{c}{\textbf{MSD}} \\
     & \textbf{E2L} & \textbf{L2E} & \textbf{E2L} & \textbf{L2E} \\ \hline
    OpenNMT+PT & .2695 & - & .2204 & - \\ 
    UNTS & .3115 & - & .3313 & - \\ \hline
    ControlledGen & .3187 & .2856 & .2170 & .1636 \\ 
    DeleteAndRetrieve & .1983 & .1684 & .3378 & .3345 \\ 
    StyleTransformer & .3189 & .2933 & .3541 & .3411 \\ \bottomrule
  \end{tabular}
  \caption{Performance using SARI.}
	\label{tab:sari}
\end{table}

\begin{table*}[htp]
  \tablesepremove
  \centering
  \small
	\begin{tabular}{c|p{12.5cm}}
    \toprule
    \textbf{Expertise input} & Prostate cancer usually progresses slowly and rarely causes symptoms until advanced. \\
    \textbf{OpenNMT+PT} & Prostate cancer usually {\color{red} \underline{goes}} slowly and rarely causes symptoms until advanced. \\ 
    \textbf{UNTS} & Prostate cancer usually {\color{red} \underline{goes}} slowly and rarely causes symptoms until advanced. \\ 
    \textbf{ControlledGen} & Prostate cancer usually progresses slowly and rarely causes symptoms until advanced. \\ 
    \textbf{DeleteAndRetrieve} & prostate cancer usually {\color{red} \underline{begins to develop until symptoms appear}}. \\ 
    \textbf{StyleTransformer} & Prostate cancer usually progresses slowly and rarely causes symptoms until advanced. \\ \hline
    \textbf{Laymen Gold} & Prostate cancer usually {\color{blue} \underline{causes no symptoms until it reaches an advanced stage}}. \\ 
    \midrule\midrule
    \textbf{Expertise input} & Cystic lung disease and recurrent spontaneous pneumothorax may occur. These disorders can cause pain and shortness of breath. \\ 
    \textbf{OpenNMT+PT} & Cystic lung disease can cause pain and shortness of breath. \\
    \textbf{UNTS} & \  {\color{red} \underline{lung}} lung disease and {\color{red} \underline{roughly something}} pneumothorax may occur. \\
    \textbf{ControlledGen} & Cystic lung disease and recurrent spontaneous pneumothorax may occur. These disorders can cause pain and shortness of breath. \\ 
    \textbf{DeleteAndRetrieve} & \ {\color{red} \underline{ear skin disease in the lungs and the lungs}} may occur {\color{red} in other disorders} and may cause {\color{red} \underline{chest}} pain and shortness of breath. \\ 
    \textbf{StyleTransformer} & Cystic lung disease and {\color{red} \underline{exposed spontaneous}} pneumothorax may occur. \\ 
    \textbf{Laymen Gold} & \  {\color{blue} \underline{Air-filled sacs (cysts) may develop in the lungs. The cysts may rupture, bringing air into the space} \underline{that surrounds the lungs (pneumothorax)}}. These disorders can cause pain and shortness of breath. \\
		\bottomrule
  \end{tabular}
  \vspace{-0.2cm}
  \caption{Examples of model outputs. Red/blue words with underlines highlight model/expected modifications.}
	\label{tab:qua_ana}
\end{table*}

Table~\ref{tab:sari} presents the performance based on the TS evaluation metric, SARI. We utilize the Python package\footnote{\url{https://github.com/cocoxu/simplification}} and follow the settings in the original paper. Surprisingly, SARI on MSD presents a relatively comprehensive evaluation that is consistent with the above analysis as well as our intuition. ControlledGen and OpenNMT+PT are ranked lower since they tend to simply repeat the input. DeleteAndRetrieve and UNTS are ranked in the middle due to the accurate style transfer but poor content preservation. StyleTransformer is ranked highest as it performs stably in Table~\ref{tab:perf} and Figure~\ref{fig:abl_concept},~\ref{fig:abl_ed},~\ref{fig:abl_type}. This inspires us to further investigate automatic evaluation metrics based on TS studies, which is our ongoing work. Even so, we still recommend necessary human evaluation in the current stage.

\subsection{Case Study}
Table~\ref{tab:qua_ana} presents two examples of transferred sentences. In the first example, both OpenNMT+PT and UNTS make lexical changes: replacing \textit{progresses} with \textit{goes}. DeleteAndRetrieve transfers style successfully but also changes the content slightly. The other two output the original expert sentence, that is the reason why they achieve higher BLEU (also PPL) but fails in Acc. Manipulation method (i.e., DeleteAndRetrieve) is more progressive in changing the style, but disentanglement method, ControlledGen, prefers to stay the same.

The second example shows structural modifications. We can see that the supervised OpenNMT+PT simply deletes the complex terminologies \textit{recurrent spontaneous pneumothorax}, but the output sentence can be deemed correct. ControlledGen still outputs the original input sentence, and the other three fail by either simply cutting the long sentence off, or changing the complex words randomly. Besides, all of the above models still perform much worse than human, which motivates research into better models.

\subsection{Discussion}
\label{sec:disc}
We have two observations from the aspects of model and evaluation. For models, there is a huge gap between all of the above models and human references. MSD is indeed challenging to conduct language modifications considering both knowledge and structures. Most of the time, these models basically output the original sentences without any modifications, or simply cut off the complex long sentence. Therefore, it is exciting to combine the techniques in TS, such as syntactic revisions including sentence splitting and lexical substitutions, with the techniques in ST: style and content disentanglement or the unsupervised idea of alleviating the lack of parallel training data.

For evaluation, human checking is necessary in the current stage, even though SARI seems to offer a good start for automatic evaluation. Based on our observations, it is actually easy to fool the three ST metrics simultaneously via a trick: \textit{output sentences by adding style-related words before the original inputs}. This is demonstrated by a variant of ControlledGen. We incorporate into the generator an extra knowledge encoder, which encodes the definition of concepts in each sentence (as mentioned in Section~\ref{sec:data_build}). Surprisingly, such a simple model achieves a very high style accuracy (over 90\%) and good BLEU scores (around 20). But the model does not succeed in the style transfer task, and simply learns to add the word \textit{doctors} into layman sentences while almost keeping the other words unchanged; and adding the word \textit{eg} into the expertise sentences. Thus, it achieves good performance on all of the three ST measures, but makes little useful modifications.

\section{Conclusion}
We proposed a practical task of expertise style transfer and constructed a high-quality dataset, MSD. It is of high quality and also challenging due to the presence of knowledge gap and the need of structural modifications. We established benchmark performance of five SOTA models. The results shown a significant gap between machine and human performance. Our further discussion analyzed the challenges of existing metrics. 

In the future, we are interested in injecting knowledge into text representation learning~\cite{cao2017bridge,cao2018joint} for deeply understanding expert language, and will help to generate knowledge-enhanced questions~\cite{pan2019recent} for laymen.

\section*{Acknowledgments}

This research is supported by the National Research Foundation, Singapore under its International Research Centres in Singapore Funding Initiative. Any opinions, findings and conclusions or recommendations expressed in this material are those of the author(s) and do not reflect the views of National Research Foundation, Singapore.

\bibliography{ms}
\bibliographystyle{acl_natbib}



\end{document}